\documentclass{article}[10pt]

\usepackage{epsfig}
\usepackage{url}

\newcommand{\bd}{\begin{document}}
\newcommand{\ed}{\end{document}}
\newcommand{\bc}{\begin{center}}
\newcommand{\ec}{\end{center}}
\newcommand{\vs}{\vspace}
\newcommand{\hs}{\hspace}
\newcommand{\beq}{\begin{equation}}
\newcommand{\eeq}{\end{equation}}
\newcommand{\beqs}{\begin{eqn*}}
\newcommand{\eeqs}{\end{eqn*}}
\newcommand{\bq}{\begin{quote}}
\newcommand{\eq}{\end{quote}}
\newcommand{\lb}{\linebreak}
\newcommand{\mb}{\makebox}
\newcommand{\fb}{\framebox}
\newcommand{\mc}{\multicolumn}
\newcommand{\ben}{\begin{enumerate}}
\newcommand{\een}{\end{enumerate}}
\newcommand{\bit}{\begin{itemize}}
\newcommand{\eit}{\end{itemize}}
\newcommand{\ov}{\overline}
\newcommand{\un}{\underline}
\newcommand{\lt}{\left}
\newcommand{\rt}{\right}
\newcommand{\ba}{\begin{array}}
\newcommand{\ea}{\end{array}}
\newcommand{\beqa}{\begin{eqnarray}}
\newcommand{\eeqa}{\end{eqnarray}}
\newcommand{\beqas}{\begin{eqnarray*}}
\newcommand{\eeqas}{\end{eqnarray*}}
\newcommand{\bfg}{\begin{figure}}
\newcommand{\efg}{\end{figure}}
\newcommand{\pad}{\partial}
\newcommand{\nn}{\nonumber}
\newcommand{\la}{\leftarrow}
\newcommand{\ra}{\rightarrow}
\newcommand{\lgla}{\longleftarrow}
\newcommand{\lgra}{\longrightarrow}
\newcommand{\La}{\Leftarrow}
\newcommand{\Ra}{\Rightarrow}
\newcommand{\Lra}{\Leftrightarrow}
\newcommand{\Lgla}{\Longleftarrow}
\newcommand{\Lgra}{\Longrightarrow}
\renewcommand{\a}{\alpha}
\renewcommand{\b}{\beta}
\newcommand{\g}{\gamma}
\newcommand{\G}{\Gamma}
\renewcommand{\d}{\delta}
\newcommand{\D}{\Delta}
\newcommand{\e}{\epsilon}
\newcommand{\eps}{\epsilon}
\newcommand{\s}{\sigma}
\renewcommand{\l}{\lamda}
\newcommand{\m}{\mu}
\newcommand{\n}{\nu}
\renewcommand{\S}{\Sigma}
\newcommand{\p}{\pi}
\newcommand{\om}{\omega}
\newcommand{\Om}{\Omega}
\newcommand{\tri}{\triangle}
\newcommand{\ti}{\times}
\newcommand{\f}{\frac}
\newcommand{\ds}{\displaystyle}
\newcommand{\bm}[1]{\mb{{\boldmath $#1$}}}
\newcommand{\alter}[2]{\lt\{ \ba{ll}#1 \\ #2 \ea \rt.}
\newcommand{\alt}[4]{\lt\{ \ba{ll}#1 & \mb{if \, \,}#2 \\ #3 & \mb{}#4 \ea 
	\rt.}
\newcommand{\altn}[4]{\lt\{ \ba{rl}#1 & \mb{if \, \,}#2 \\ #3 & \mb{}#4 \ea 
	\rt.}
\newcommand{\altif}[4]{\lt\{ \ba{ll}#1 & \mb{if \, \,}#2 \\ #3 & 
\mb{if \, \,}#4 \ea \rt.}
\newcommand{\altnif}[4]{\lt\{ \ba{rl}#1 & \mb{if \, \,}#2 \\ #3 & 
\mb{if \, \,}#4 \ea \rt.}
\newcounter{algc}
\newcounter{romc}
\newcounter{Alphc}
\newcommand{\bl}{\begin{list}{{\it Step} ~\arabic{algc}~:} {\usecounter{algc}
                \setlength{\topsep}{0pt} \setlength{\itemsep}{0pt}}}
\newcommand{\el}{\end{list}}
\newcommand{\blr}{\begin{list}{~\roman{romc}~:} {\usecounter{romc}
                \setlength{\topsep}{0pt} \setlength{\itemsep}{0pt}}}
\newcommand{\elr}{\end{list}}
\newcommand{\bla}{\begin{list}{~\Alph{Alphc}~:} {\usecounter{Alphc}
                \setlength{\topsep}{0pt} \setlength{\itemsep}{0pt}}}
\newcommand{\ela}{\end{list}}
\newtheorem{theorem}{Theorem}

\bd 

\bibliographystyle{plain}
\setlength{\topmargin}{-0.2cm}
\setlength{\leftmargin}{-2.0cm}
\setlength{\parskip}{1.0pc}
\setlength{\parindent}{0.0pt}


\Large{\bf ASOC: An Adaptive Parameter-free Stochastic Optimization Techinique for Continuous Variables}

{\bf \em Jayanta Basak\\
Advanced Technology Group, NetApp,\\
Bangalore, India.} \\
{\small basak@netapp.com,basakjayanta@yahoo.com}


{\bf Abstract}\\
Stochastic optimization is an important task in many optimization problems where the tasks are not
expressible as convex optimization problems. In the case of non-convex optimization problems, various different
stochastic algorithms like simulated annealing, evolutionary algorithms, and tabu search are available. Most of these
algorithms require user-defined parameters specific to the problem in order to find out the optimal solution. Moreover,
in many situations, iterative fine-tunings are required for the user-defined parameters, and therefore these algorithms
cannot adapt if the search space and the optima changes over time. In this paper we propose an \underline{a}daptive 
parameter-free \underline{s}tochastic \underline{o}ptimization technique for \underline{c}ontinuous random variables
called ASOC.




\section{Introduction}

Stochastic optimization \cite{Wikipedia14a} is the task of optimizing certain objective functional by generating 
and using stochastic random
variables. Usually the stochastic optimization is an iterative process of generating random variables that progressively
finds out the minima or the maxima of the objective functional. Stochastic optimization is usually applied in the
non-convex functional spaces where the usual deterministic optimization such as linear or quadratic programming or their
variants cannot be used. Stochastic optimization is performed in discrete spaces such as generalized hill climbing 
\cite{Johnson02},
and continuous spaces \cite{Bennett06}. In this article, we focus only on the stochastic optimization task in 
the continuous domain.

Stochastic optimization in continuous domain includes a large number of different algorithms that include stochastic 
gradient descent \cite{Kiwiel03},
simulated annealing \cite{Kirkpatrick83,Fox95,Faigle91,Fielding00,Geman84}, evolutionary algorithms 
\cite{Goldberg89,Fogel94}, 
tabu search \cite{Glover94,Fox93}, and many others. Stochastic gradient descent 
and quasi-Newton techniques usually find out a local optima in the search space. On the other hand, simulated annealing
can find out the global optima with a proper temperature schedule \cite{Fox95}. Evolutionary algorithms are also proven to
obtain the global optima under certain conditions \cite{Fogel94}. However, most of the existing techniques require specification of
user-defined parameters. For example, simulated annealing performance is highly dependent on the cooling schedule. 
Evolutionary algorithms depend on the crossover and mutation probabilities that are defined by the user. Secondly,
most of the stochastics search techniques operate with tunable parameters. For example, in simulated annealing, the 
temperature is gradually reduced with certain cooling schedule. In evolutionary algorithms also, the crossover and
mutation probabilities are usually reduced over iterations. In other words, these algorithms are mostly not adaptive.
By adaptivity of an algorithm we mean, if the objective functional changes over time, the algorithm will be able to
follow the new optimal points according to the changed search space structure. If the user-defined parameters are reduced
gradually, the algorithm converges to the optimal point but loses the capability of adjusting the solution space to
the changing search space structure if the objective functional changes.

In this paper, we propose a parameter-free \underline{a}daptive \underline{s}tochastic \underline{o}ptimization algorithm 
for \underline{c}ontinuous random variables (ASOC) that is not only independent of
the choice of any user-defined parameter but is also able to adapt to the changes in the serach space structure for
a changing objective functional. We derive the idea of optimization from the generative models in pattern classification 
\cite{Tu07}.
First we consider a sample pool and obtain their corresponding functional values. We then define ordered pairs of the
samples in such a way that if a sample has less functional value than that of the next sample in the ordered pair then
it belongs to a particular class. We then iteratively generate ordered pairs from this class such that  the first sample
in the ordered pair has less functional value than the second sample in the ordered pair. Thus we iteratively generate
samples as obtained from the generated ordered pairs that progressively reduces the functional value. As the process
converges i.e., there is no more decrease in the functional value, we obtain the minima of the optimization function. An
analogous process can be followed if the task is to maximize the objective function. ASOC has a similarity with the
stochastic gradient descent where a sample is updated based on the local gradient of the objective function 
\cite{Kiwiel03}. However,
we never compute the gradient of the objective function explicitly. In other words, we are not constrained by the fact
that the objective function need to be locally differentiable. ASOC can be applied to any stochastic optimization for continuous
variables even if the function is not expressible in a mathematical form but can be computed using the sample values.
In the literature of Bayesian optimization \cite{Pelikan00}, a similar approach is followed, however, the Bayesian 
optimization techniques
do not use the concept of generative models of the ordered pairs to minimize or maximize the functional values.

\section{Problem Formulation}

\subsection{Representation}

Let the optimization problem be finding out an ${\bf x_{opt}}$,
\begin{equation}
{\bf x_{opt}} = argmin_{\bf x}\{ f({\bf x}) \}
\end{equation}
subject to ${\bf x} \in {\cal D} \subset {\cal R}^{n}$ such that the task is to find out the minima 
of the function $f({\bf x})$. Here $f({\bf x})$ is not necessarily expressible in parametric form and not necessarily 
a smooth function. In practice, several such optimization tasks exist where it is extremely difficult to express a
suitable functional form of the optimization problem mathematically. In this paper, we do not assume any form of 
the function. 
The optimization problem is such that for any given $n$-dimensional vector 
${\bf x} \in {\cal D} \subset {\cal R}^n$, the objective can be evaluated. 

The generic representation structure of the proposed algorithm is analogous to that of the evolutionary algorithms. 
Here we maintain a pool of $N$ vectors $X = \{{\bf x_1}, {\bf x_2}, \cdots, {\bf x_N}\}$ and their corresponding 
objective values 
$f(X) = \{ f({\bf x_1}), f({\bf x_2}), \cdots, f({\bf x_N})\}$. The algorithm procees iteratively, and at 
every iteration it 
generates a new pool of candidate vectors $X_*$. The algorithm then finds out a set of best fitting candidate vectors,
as evaluated by the objective function,
from the $X\bigcup X_*$. Next, the entire process is repeated until there is no more change in the best fitting
solution. The strategy of generating new candidate vectors is derived from the idea of generative models
in pattern classification task \cite{Tu07} where
we define synthetic class structures consisting of ordered pair of samples. We then generate new samples from this class
structure such that a new sample is randomly drawn that is expected to be better than the best in the
existing pool of samples.

\subsection{Optimization as a Generative Model}

As mentioned before, the pool of $N$ candidate vectors is represented as $X = \{{\bf x_1}, {\bf x_2}, \cdots, {\bf x_N}\}$. 
Without loss of generality, let us assume that the pool of vectors be sorted as 
${\bf x_1} \prec {\bf x_2} \prec \cdots \prec {\bf x_N}$ according to their objective 
values such that in $f(X)$, 
\begin{equation}
f({\bf x_1}) \leq f({\bf x_2}) \leq \cdots \leq f({\bf x_N})
\end{equation}
where $f({\bf x})$ denote the objective functional value of the vector ${\bf x}$. With this representation, 
we transform the problem into a space of ordered pair of vectors ${\bf y_{ij}} = [{\bf x_i}, {\bf x_j}]^{'}$ where
$'$ indicates transpose. 
In other words, let the vector notation of the $n$-dimensional vector ${\bf x_i}$ be given as 
${\bf x_i} = [x_{i1},x_{i2},\cdots,x_{in}]^{'}$. The concatenated vector ${\bf y_{ij}}$ is then given as
\begin{equation}
{\bf y_{ij}} = [x_{i1},x_{i2},\cdots,x_{in},x_{j1},x_{j2},\cdots,x_{jn}]^{'}
\end{equation}
We therefore obtain $N(N-1)$ such ordered pair of vectors for all $i$, $j$, $i \neq j$. 
We partition these concatenated vectors into two classes namely $\Omega_1$ and $\Omega_2$ each 
containing $N(N-1)/2$ concatenated samples such that 
\begin{equation}
{\bf y_{ij}} \in \begin{array}{ll} \Omega_1 & \mbox{if ${\bf x_i} \prec {\bf x_j}$} \\ \Omega_2 & \mbox{otherwise} \end{array}
\label{eq:yij}
\end{equation}
Once we obtain such a partition, the class structures of $\Omega_1$ and $\Omega_2$ are defined by the pool vectors 
subject to certain density estimate. Once the class structure is defined, the next task is to obtain one candidate 
vector ${\bf x_*}$ such that
\begin{equation}
{\bf y_{*i}} \in \Omega_1
\label{eq:y*i}
\end{equation}
for all ${\bf x_i} \in X$. In other words, we need to find out one candidate 
vector ${\bf x_*}$ which is better than the existing pool vectors in terms of the objective values. 
It is equivalent to finding out one ${\bf x_*}$ that is better than the best pool vector such that
\begin{equation}
{\bf y_{*1}} \in \Omega_1
\label{eq:y*1}
\end{equation}
with a sorted pool $X$. 

In order to find ${\bf x_*}$, we use the concept of conditional distribution of $X_*$ conditioned on $X$ such that 
Equation (\ref{eq:y*1})
is satisfied. The distribution of $Y \in \Omega_1$ is approximated as normal ${\cal N}({\bf \mu},{\bf \Sigma})$ such that
\begin{equation}
{\bf \mu} = [{\bf \mu_1}, {\bf \mu_2}]^{'}
\label{eq:mu}
\end{equation}
and
\begin{equation}
{\bf \Sigma} = \left[ \begin{array}{ll} {\bf \Sigma_{11}} & {\bf \Sigma_{12}} \\ {\bf \Sigma_{21}} & {\bf \Sigma_{22}} \end{array} \right]
\label{eq:Sigma}
\end{equation}
where
${\bf \mu}$ and ${\bf \Sigma}$ are determined from all samples in $\Omega_1$. Then the distribution of $X_*$ with 
the condition 
$X= {\bf x_1}$ is given as ${\cal N}({\bf \hat{\mu}},{\bf \hat{\Sigma}})$ where
\begin{equation}
{\bf \hat{\mu}} = {\bf \mu_1} + {\bf \Sigma_{12}\Sigma_{22}^{-1}(x1 - \mu_2)}
\label{eq:hatmu}
\end{equation}
and ${\bf \hat{\Sigma}}$ is given as (Schur complement)\cite{Wikipedia15}
\begin{equation}
{\bf \hat{\Sigma}} = {\bf \Sigma_{11} - \Sigma_{12}\Sigma_{22}^{-1}\Sigma_{21}}
\label{eq:hatSigma}
\end{equation}
Once we obtain the distribution of $X_*$ as ${\cal N}({\bf \hat{\mu}},{\bf \hat{\Sigma}})$, we
generate new samples from that distribution.

The new sample generation process is similar to stochastic gradient descent \cite{Kiwiel03} process except the fact 
that the new
samples are generated from the estimated target distribution instead of a deterministic point generated from the
gradient. The nature
of the target distribution depends on the previous distribution of the samples. 
In simulated annealing, the acceptance probability of an inferior solution is 
modulated by $\exp(-\Delta E/T)$ where $T$ is the temperature and $\Delta E$ is the increase in the objective functional
value of the inferior solution. As $T$ goes to zero, the acceptance probability goes to zero.
In ASOC, we guide the selection process to iteratively adapt the new solution towards the minima. In our case, there is no
temperature schedule or cooling process as used in the simulated annealing. Our technique is completely adaptive and 
depends on the pool of samples.
Even if the functional value changes, the technique automatically adapts the samples to select the new optima. 

We start
with a randomly generated sample pool in {\cal D}. Let $S$ be a sample pool having $2N$ samples. We first sort the $S$ 
in ascending order
according to the functional values of the samples, and select
the top $N$ samples from that pool. Let this sample pool be $X$ having $N$ samples. We then compute 
$({\bf \hat{\mu}},{\bf \hat{\Sigma}})$
from these sample pool $X$. Next we draw $N$ samples randomly in {\cal D} using ${\cal N}(\hat{\mu},\hat{\Sigma})$. 
Let this sample
pool be $X_*$. We then have the sample set $S = X \bigcup X_*$, and again sort $S$ in ascending order of the 
functional
values and repeat the entire process to generate new set of samples. We iteratively generate the new samples until there
is no significant change in the best solution. Note that the samples in $X_*$ may be inferior to the best sample in $X$
and in that case the best sample in $X$ will automatically move to the next iteration. In other words, we always follow
the elitist selection mechanism unlike the simulated annealing.

\subsection{Overall Algorithm}

We summarize the overall algorithm in this section.

{\it Problem:} Find the minima of a given objective function $f(.)$ in the $n$-dimensional continuous space such that
\begin{equation}
{\bf x_{opt}} = argmin_{\bf x}\{ f({\bf x}) \}
\end{equation}
subject to ${\bf x} \in {\cal D} \subset {\cal R}^{n}$. The objective function need not be continuous and differentiable.

\begin{itemize}
\item[Step 1:] Randomly initialize a sample pool $S$ with $2N$ samples in the $n$-dimensional space such that each sample is in ${\cal D}$.
\item[Step 2:] Sort the sample pool $S$ in the ascending order of the functional values and choose top $N$ samples
from the sorted pool to construct the sample set $X$.
\item[Step 3:] Rank order the samples in $X$ such that for any $i < j$, $f({\bf x_i}) \leq f({\bf x_j})$. 
Construct the class $\Omega_1$
as described in Equation (\ref{eq:yij}). Estimate $({\bf \mu},{\bf \Sigma})$ of class $\Omega_1$ as described in 
Equation (\ref{eq:mu}) and (\ref{eq:Sigma}).
\item[Step 4:] Estimate the target distribution ${\cal N}({\bf \hat{\mu}}, {\bf \hat{\Sigma}})$ as described in 
Equaiton (\ref{eq:hatmu}) and (\ref{eq:hatSigma}).
\item[Step 5:] Randomly draw $N$ samples from the target distribution ${\cal N}({\bf \hat{\mu}}, {\bf \hat{\Sigma}})$ and 
constrain the samples
such that the sample set $X_{*} \in {\cal D}$. Construct the set $S = X \bigcup X_{*}$.
\item[Step 6:] Repeat the process from {\it Step 2} until some stopping criteria is satisfied.

\end{itemize}

In ASOC, if  
$\hat{\Sigma} \rightarrow {\bf 0}$ then any new sample vector is not generated. However, we do not omit the inferior
samples from a sample pool but they become iteratively better. Thus even if there is no change in the best sample
in the sample pool, the other samples may get iteratively better.

One of the major advantage of the proposed search algorithm is that it is totally free from any user defined
parameter. The state-of-the-art stochastic search algorithms such as the class of simulated annealing and genetic 
algorithms highly depend on user-defined parameters. For example, in simulated annealing, the search process is
guided by an artificial cooling schedule defined by temperature. The schedule of decreasing the temperture is
decided beforehand. Similarly, in evolutionary algorithms, the performance depends on the crossover and
mutation probabilities and these probability values are user-defined.

\section{Experimental Results}

There exists a large number of benchmark functions in the literature \cite{Jamil13} for testing the effectiveness of
stochastic optimization algorithms. A subset of these functions is available in \cite{Wikipedia14b}. We used the 
same subset of
functions as in \cite{Wikipedia14b} for testing the effectiveness of ASOC. Table \ref{tbl:function} enlists the 
functions that we used 
in our experiments.
We demonstrate the effectiveness of ASOC in optimizing these functions and compare ASOC with simulated annealing
and genetic algorithms using the same set of functions.

We have implemented the ASOC in Matlab in the Windows XP environment. We have chosen a population size (N) = 30 and 
observed the convergence properties of the ASOC
algorithm for 2000 generations. As a comparison, we optimized the functions using both simulated annealing 
and genetic algorithm for continuous variables. In simlated annelaing, we iterated for 2000 iterations and in each 
iteration we generated samples randomly with a constant temperature for 50 times. We reduced the temperature following
a logarithmic schedule over 2000 iterations. For the genetic algorithm, we used elitist model where the best chromosome 
is always passed into the next generation. 


In Table \ref{tbl:performance}, we show the effectiveness of ASOC along with SA and GA for 100, 500, and 2000 iterations 
respectively.
In the implementation of SA, the temperature has been reduced according to the number of iterations. For a smaller
number of iterations, the temperature is reduced quickly and for a large number of iterations, temperature is
reduced rather slowly.

From Table \ref{tbl:performance}, we observe that GA and ASOC can obtain the optimal points in most of the cases. 
For Easom function,
none of the techniques are successful in obtaining the minimal point. For Rosenbrock function, we observe that ASOC
outperforms GA for a dimensionality equal to 3. For the same function, simulated annealing did not converge. 

\begin{figure}[htbp]
\centering
\epsfig{file=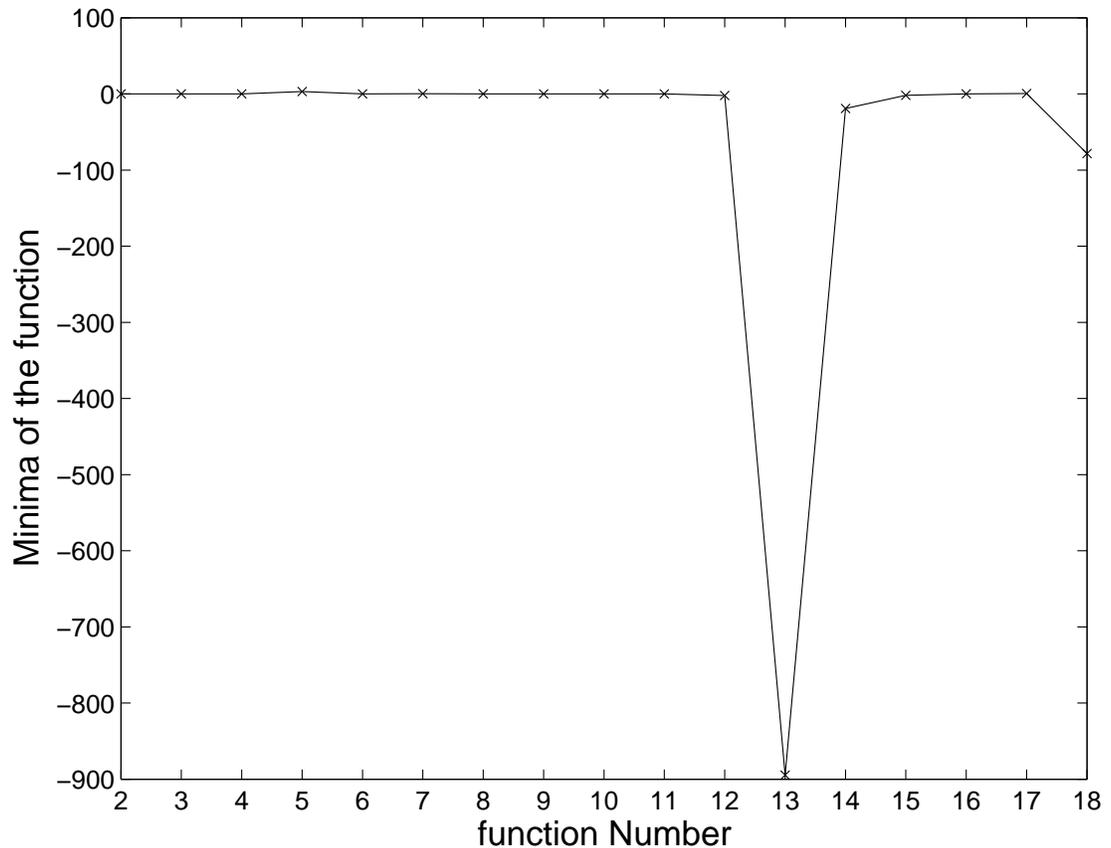, width=\columnwidth}
\caption{Minima obtained adaptively as the function changes}
\label{fig:adaptive}
\end{figure}

In simulated annealing, the temperature is reduced to obtain the global optima. However, if the nature of the
optimization function changes, then SA will not be able to adapt to the new situation and find the new optima. On the
other hand, ASOC is practically parameter-free optimization technique and it continues to generate new samples
in the vicinity of the optima once it has converged. If the nature of the optimization function changes then it
gracefully switches over to the new optima location and adapts the solution space. In order to show the effectiveness
of ASOC to adapt to the new situation, we change the function from function number 2 to 18 (as in 
Table \ref{tbl:function}) and
run ASOC for each function for 2000 iterations without reinitializing the samples. It is as if once the algorithm has
converged, a new optima appears. We did not consider the function number 1 (Table \ref{tbl:function}) because 
function 1 and function 2
has the same optima locations. Figure \ref{fig:adaptive} illustrates how the optima changes as the function changes. 
We observe that
ASOC is indeed able to follow the changing pattern of the optimization problem.

\section{Discussion}

We have presented a new adaptive parameter-free stochastic optimization technique called ASOC. We have demonstrated that
ASOC can find the optimal solution on certain benchmark problems. Simulated annealing converges to the global optima
with suitably chosen cooling schedule \cite{Faigle91,Granville94}. Evolutionary algorithms are also globally convergent 
under certain conditions \cite{Fogel94}.
The convergence properties of ASOC require further analysis. A possible approach towards proving the convergence
under a generalized framework of such optimization algorithms is provided in \cite{Zhang04,Dawid94}.

We generate new samples considering the class of ordered pair of samples as a single cluster and therefore we derive 
a single
mean and covariance matrix. It is possible to extend ASOC by clustering the ordered pair of samples into different clusters
and generating means and covariance matrices for each cluster separately. In this way, we will have more variability
in the generated samples that may lead to better convergence.
ASOC finds out one optimal point for a given single objective functional. It is possible to extend ASOC to find out 
pareto-optimal solutions for multi-objective functionals as a constituent future work.


\begin{table}
\caption{List of functions used to evaluate ASOC, SA, and GA.}
{\tiny
\begin{tabular}{|l|l|l|l|} \hline
Function & Mathematical Expression & Minima & Domain \\ \hline
Ackley &
$f(x,y) = -20\exp\left(-0.2\sqrt{0.5\left(x^{2}+y^{2}\right)}\right)$ 
& $f(0,0) = 0$
& $-5\le x,y \le 5$
\\
& $-\exp\left(0.5\left(\cos\left(2\pi x\right)+\cos\left(2\pi y\right)\right)\right) + e + 20$ & & \\ \hline
Sphere 
& $f(\bf{x}) = \sum_{i=1}^{n} x_{i}^{2}$ 
& $f(x_{1}, \dots, x_{n}) = f(0, \dots, 0) = 0$ 
& $-\infty \le x_{i} \le \infty$, $1 \le i \le n$ 
\\ \hline
Rosenbrock  
& $f(\bf{x}) = \sum_{i=1}^{n-1} \left[ 100 \left(x_{i+1} - x_{i}^{2}\right)^{2} + \left(x_{i} - 1\right)^{2}\right]$  
& $f_{min} =
\left \{ \begin{array}{ll}
n=2 & \rightarrow  f(1,1) = 0, \\
n=3 & \rightarrow  f(1,1,1) = 0, \\
n>3 & \rightarrow  f\left(\underbrace{1,\dots,1}_{(n) \mbox{ times}}\right) = 0 
\end{array} \right.
$
& $-\infty \le x_{i} \le \infty$, $1 \le i \le n$ 
\\ \hline
Beale 
& $f(x,y) = \left( 1.5 - x + xy \right)^{2} + \left( 2.25 - x + xy^{2}\right)^{2}$ 
& $f(3, 0.5) = 0$
& $-4.5 \le x,y \le 4.5$ 
\\ 
& $+ \left(2.625 - x+ xy^{3}\right)^{2}$ & & \\ \hline
Goldstein–Price 
& $f(x,y) = \left(1+\left(x+y+1\right)^{2}\left(19-14x+3x^{2}-14y+6xy+3y^{2}\right)\right)$ 
& $f(0, -1) = 3$
& $-2 \le x,y \le 2$
\\ 
& $\left(30+\left(2x-3y\right)^{2}\left(18-32x+12x^{2}+48y-36xy+27y^{2}\right)\right)$ & & \\ \hline
Booth
&$f(x,y) = \left( x + 2y -7\right)^{2} + \left(2x +y - 5\right)^{2}$
&$f(1,3) = 0$
&$-10 \le x,y \le 10$
\\ \hline
Bukin N.6
& $f(x,y) = 100\sqrt{\left|y - 0.01x^{2}\right|} + 0.01 \left|x+10 \right|.\quad$
& $f(-10,1) = 0$
& $-15\le x \le -5$, $-3\le y \le 3$
\\ \hline
Matyas 
& $f(x,y) = 0.26 \left( x^{2} + y^{2}\right) - 0.48 xy$
& $f(0,0) = 0$
& $-10\le x,y \le 10$
\\ \hline
L\'{e}vi N.13:
& $f(x,y) = \sin^{2}\left(3\pi x\right)+\left(x-1\right)^{2}\left(1+\sin^{2}\left(3\pi y\right)\right)$ 
$+\left(y-1\right)^{2}\left(1+\sin^{2}\left(2\pi y\right)\right)$
& $f(1,1) = 0$
& $-10\le x,y \le 10$
\\ 
& $+\left(y-1\right)^{2}\left(1+\sin^{2}\left(2\pi y\right)\right)$ & & \\ \hline
Three-hump camel 
& $f(x,y) = 2x^{2} - 1.05x^{4} + \frac{x^{6}}{6} + xy + y^{2}$
& $f(0,0) = 0$
& $-5\le x,y \le 5$
\\ \hline
Easom 
& $f(x,y) = -\cos \left(x\right)\cos \left(y\right) \exp\left(-\left(\left(x-\pi\right)^{2} + \left(y-\pi\right)^{2}\right)\right)$
& $f(\pi , \pi) = -1$
& $-100\le x,y \le 100$
\\ \hline
Cross-in-tray 
& $f(x,y) = -0.0001 \left( \left| \sin \left(x\right) \sin \left(y\right) \exp \left( \left|100 - \frac{\sqrt{x^{2} + y^{2}}}{\pi} \right|\right)\right| + 1 \right)^{0.1}$
& $f_{min} = \left \{ \begin{array}{ll}
      f\left(1.34941, -1.34941\right) & = -2.06261 \\
      f\left(1.34941,  1.34941\right) & = -2.06261 \\
      f\left(-1.34941, 1.34941\right) & = -2.06261 \\
      f\left(-1.34941,-1.34941\right) & = -2.06261 
\end{array} \right.
$
& $-10\le x,y \le 10$
\\ \hline
Eggholder 
& $f(x,y) = - \left(y+47\right) \sin \left(\sqrt{\left|y + \frac{x}{2}+47\right|}\right) - x \sin \left(\sqrt{\left|x - \left(y + 47 \right)\right|}\right)$
& $f(512, 404.2319) = -959.6407$
& $-512\le x,y \le 512$
\\ \hline
Holder table 
& $f(x,y) = - \left|\sin \left(x\right) \cos \left(y\right) \exp \left(\left|1 - \frac{\sqrt{x^{2} + y^{2}}}{\pi} \right|\right)\right|$ 
& $f_{min} = \left\{ \begin{array}{ll}
      f\left(8.05502,  9.66459\right) & = -19.2085 \\
      f\left(-8.05502,  9.66459\right) & = -19.2085 \\
      f\left(8.05502,-9.66459\right) & = -19.2085 \\
      f\left(-8.05502,-9.66459\right) & = -19.2085
\end{array} \right.
$ 
& $-10\le x,y \le 10$
\\ \hline
McCormick 
& $f(x,y) = \sin \left(x+y\right) + \left(x-y\right)^{2} - 1.5x + 2.5y + 1$
& $f(-0.54719,-1.54719) = -1.9133$
& $-1.5\le x \le 4$, $-3\le y \le 4$
\\ \hline
Schaffer N. 2
& $f(x,y) = 0.5 + \frac{\sin^{2}\left(x^{2} - y^{2}\right) - 0.5}{\left(1 + 0.001\left(x^{2} + y^{2}\right) \right)^{2}}$
& $f(0, 0) = 0$
& $-100\le x,y \le 100$
\\ \hline
Schaffer N. 4
& $f(x,y) = 0.5 + \frac{\cos^{2}\left(\sin \left( \left|x^{2} - y^{2}\right|\right)\right) - 0.5}{\left(1 + 0.001\left(x^{2} + y^{2}\right) \right)^{2}}$
& $f(0,1.25313) = 0.292579$ 
& $-100\le x,y \le 100$
\\ \hline
Styblinski–Tang function:
& $f(\bf{x}) = \frac{\sum_{i=1}^{n} x_{i}^{4} - 16x_{i}^{2} + 5x_{i}}{2}$
& $f\left(\underbrace{-2.903534, \ldots, -2.903534}_{(n) \mbox{ times}} \right) = -39.16599n$
& $-5\le x_{i} \le 5</, 1\le i \le n$
\\ \hline
\end{tabular}
}
\label{tbl:function}
\end{table}

\begin{table}[htb!]
\caption{Performance of ASOC, SA,  and GA.}
{\footnotesize
\begin{tabular}{|l|l|l|l|l|l|l|l|l|l|l|} \hline
Function & True Minima & \multicolumn{9}{c|}{Functional Minima Obtained} \\ \cline{3-11}
& & \multicolumn{3}{c|}{SA(number of Iter)} & \multicolumn{3}{c|}{GA(number of Iter)} & \multicolumn{3}{c|}{ASOC(number of Iter)} \\ \cline{3-11}
& & 100 & 500 & 2000 & 100 & 500 & 2000 & 100 & 500 & 2000 \\ \hline
Ackley & 0 & 0.3483 & 0.5114 & 0.1249 & 0.04712 & 0.01396 & 0.00036 & 0.06824 &  0.008 & 0.008
\\ \hline
Sphere & 0 & 0.9262 & 0.4082 & 0.327 & 0.01725 & 0.0057 & 0.00005 &  0.0111 & 0.0103 & 0.007 \\
(n=10) & & & & & & & & & & 
\\ \hline
Rosenbrock & 0 & - & - & - & 48.2698 & 11.4849 & 5.8252 & 30.6698 & 1.0576 & 1.0576 \\
(n = 3) & & & & & & & & & & 
\\ \hline
Beale & 0 & 0.8756 & 0.0121 & 0.0024 & 0.01988 & 0.0178 & 0.0155 & 0.0002 & 0 & 0
\\ \hline
Goldstein–Price & 3 & 3.241 & 3.1450 & 3.012 & 3.0318 & 3.0004 & 3.0001 & 3.0062 & 3.0022 & 3.001
\\ \hline
Booth & 0 &  5.6841 & 0.0243 & 0.0017 & 0.0804 & 0.0141 & 0.0007 & 0.0041 & 0.0004 & 0.00003 
\\ \hline
Bukin N.6 & 0 & 0.4879 & 3.7391 & 3.4592 & 0.1132 & 0.1132 & 0.1132 & 1.2977 & 0.2921 & 0.2701
\\ \hline
Matyas & 0 & 1.6888 & 0.0366 & 0.0008 & 0.0016 & 0.0009 & 0.00065 & 0.00007 & 0.00005 & 0.000006
\\ \hline
L\'{e}vi N.13 & 0 & 1.4343 & 0.0235 & 0.0192 & 0.0004 & 0 & 0 & 0.01947 & 0.0015 & 0.00001
\\ \hline
Three-hump camel & 0 & 0.8122 & 0.0046 & 0.0014 & 0.2987 & 0.0002 & 0.000001 & 0.00015 & 0.000004 &  0.000004
\\ \hline
Easom & -1 & 0 & 0 & 0 & -0.00899 & -0.009 & -0.009 & -0.0088 & -0.0089 & -0.009
\\ \hline
Cross-in-tray & -2.06261 & -1.2934 & -2.0209 & -2.0602 & -2.06261 & -2.06261 & -2.06261 & -2.06261 &  -2.06261 & -2.06261
\\ \hline
Eggholder &  -959.6407& -357.904 & -282.028 & -443.425 & -894.519 & -894.568 & -933.393 & -959.64 & -959.641 & -959.641
\\ \hline
Holder table & -19.2085 & -18.5043 & -18.7484 & -19.1858 & -19.2074 & -19.2085 & -19.2085 & -19.2016 & -19.2073 & -19.2075
\\ \hline
McCormick & -1.9133 & -1.89 & -1.9032 & -1.913 & -1.913 & -1.9132 & -1.9132 & -1.9131 & -1.9132 & -1.9132
\\ \hline
Schaffer N. 2 & 0 & 0.4388 & 0.3396 & 0.1894 & 0.0505 & 0.0091 & 0.0046 & 0.0006 & 0.0005 & 0.000001
\\ \hline
Schaffer N. 4 & 0.292579 & 0.5038 & 0.5002 & 0.5003 & 0.5001 & 0.5001 & 0.5001 & 0.500009 & 0.500009 & 0.500009
\\ \hline
Styblinski–Tang & -78.332 & -64.141 & -64.177 & -64.189 & -78.332 & -78.332 & -78.332 & -78.33 & -78.331 & -78.331 \\
(n=2) & & & & & & & & & & 
\\ \hline
\end{tabular}
}
\label{tbl:performance}
\end{table}

\ed